# An IoT Real-Time Biometric Authentication System Based on ECG Fiducial Extracted Features Using Discrete Cosine Transform

Ahmed F. Hussein
Biomedical Eng. Dept.
AlNahrain University
Baghdad, Iraq
a.f.hussein@ieee.org

Abbas K. AlZubaidi
Biomedical Eng. Dept.
AlMustaqbal Univ. College
Babil, Iraq
abbas.khudair@gmail.com

Ali Al-Bayaty
Elec. & Computer Eng. Dept.
University of New Naven
CT, USA
aalba3@unh.newhaven.edu

Qais A. Habash
Biomedical Eng. Dept.
AlNahrain University
Baghdad, Iraq
qais_hmd@yahoo.com

*Abstract—* The conventional authentication technologies, like RFID tags and authentication cards/badges, suffer from different weaknesses, therefore a prompt replacement to use biometric method of authentication should be applied instead. Biometrics, such as fingerprints, voices, and ECG signals, are unique human characters that can be used for authentication processing. In this work, we present an IoT real-time authentication system based on using extracted ECG features to identify the unknown persons. The Discrete Cosine Transform (DCT) is used as an ECG feature extraction, where it has better characteristics for real-time system implementations. There are a substantial number of researches with a high accuracy of authentication, but most of them ignore the real-time capability of authenticating individuals. With the accuracy rate of 97.78% at around 1.21 seconds of processing time, the proposed system is more suitable for use in many applications that require fast and reliable authentication processing demands.

## I. Introduction

The biometric systems for individual identification are expert and smart systems in which interests and demands are rapidly increased in many sectors, such as healthcare applications, security systems, and telecommunications. In the authentication systems, an individual person presents himself/herself as a specific individual and the system checks for his/her biometric features against a profile that exists based on a specific individuals' file to find the match, where this is usually known as a One-to-One matching system. In the next stage, the system will search for the unknown individual (biometric); in this case, the system checks the presented biometric against all of the others in the dataset, which described as a One-to-Many (N) matching system [1].

Biometric-based authentication represents one of the authentication methods that truly identifies the real applicant as a particular individual than the other traditional methods, such as passwords, ID cards, and watermarks. Although some biometrics methods currently exist, these methods are not robust enough against forgery. Whereas the private biometric passes are not protected, spoofing attack shall occur either digitally or physically. For instance, our fingerprints could exist on glasses, doors, and tables; thus, it is possible to be acquired and used. Recently, many researchers [2-4] have suggested the Electrocardiogram (ECG) as a biometrics method for individuals' authentication. For many decades ago, the ECG has been used as a heart condition diagnosing and monitoring. And, it is still the fast and the noninvasive method for identifying the primary heart problems.

As the Internet of Things (IoT) deals with many numbers of things, such as devices, sensors, and relevant data, many real security trends have to be addressed. One essential characteristic of IoT is the service customization. In IoT, the current main stream of Human-to-Device communication will be gathered to Human-to-Human (H2H), Human-to-Thing (H2T), and Thing-to-Thing (T2T). The H2T channel is still crucial, as humans often trigger various services [5]. The ECG signal, as shown in Fig. 1, has a unique morphological shape due to the anatomical structure of the heart and physiological conditions. The ECG morphology consists of, but not limited to, the P, R, and T waves amplitudes, the slope information for each segment, and the temporal distance between wave boundaries [6].

And, Fig. 2 illustrates the ECG-based biometrics analysis taxonomy and its fiducial-based and non-fiducial-based methods [7].

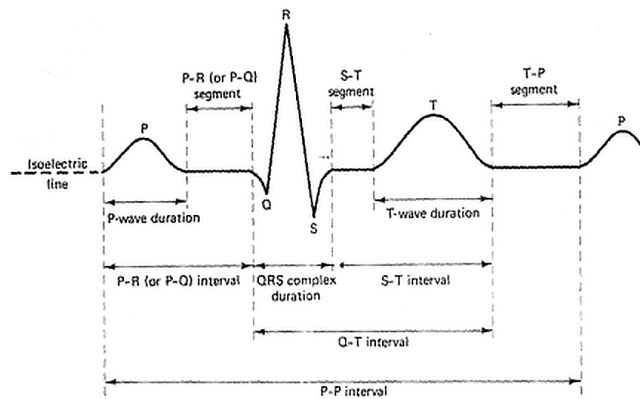

Fig.1. The morphological shape of ECG signal.

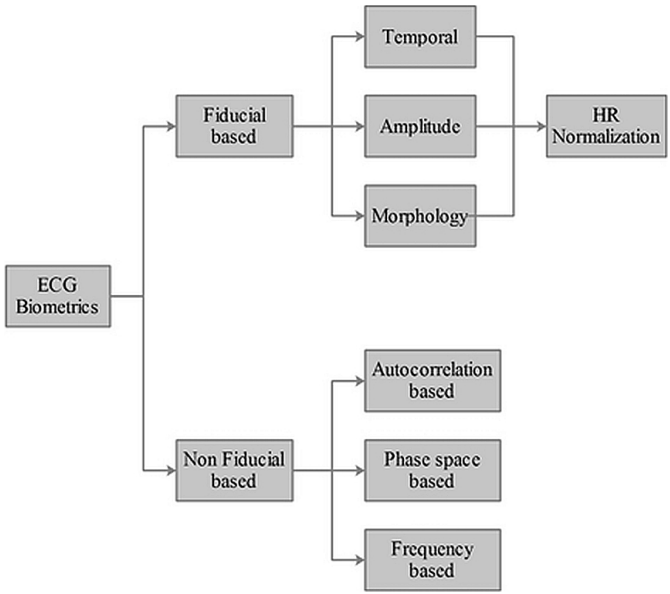

Fig. 2. The ECG-based biometrics analysis taxonomy.

Due to the fiducial marks that extracted from the ECG signal, the previous methods can be grouped into two classes: single fiducial mark and multiple fiducial mark approaches. The single fiducial mark approaches depend upon extracting the required features from the heart beats rate (the RR intervals). While the multiple fiducial mark approaches extract the features from the selected points in the ECG signal and then use it for processes classification [8].

For that, the ECG signal is valid to be used as a strong biometric authentication process according to the following characteristics, which are:

1. The ECG is a noninvasive test and easy to be measured by just placing two fingers on the sensory plates.
2. Although the general ECG morphology for each person contains the same components (P, QRS, and T), the heart's structure parameters, such as the wall thickness, position, and size, are different for each person; thus, a uniqueness ECG signal will be obtained.
3. Robustness, where the ECG-based authentication systems need limited computational processing power for recognizing individuals.

In this study, an IoT real-time authentication system that is based on the RR Discrete Cosine Transform (DCT) features is proposed. The analyzing process in this system depends on using the DCT coefficients as a feature, which is extracted from a single measured ECG beat. And, it requires capturing the ECG signal for just three seconds, where it is enough for the identification process. The system adopts *Raspberry PI 3* system as an IoT sensory node that sends the measured data for the verification process using the TCP/IP protocol. And, for validating the proposed system, the MIT-BIH ECG databases are used.

## II. STUDY BACKGROUND

Many researchers have employed the ECG for biometric authentications by using different features. Chan et al. [9] extracted the fixed-length heartbeat feature from the R-peak position. Then, they use the Wavelet coefficients with threshold technique according to the correlation for the authentication process. Kang et al. [10] presented cross-correlation templates for system identification during the authentication and the registration stages, and the proposed algorithms were implemented for wearable device verification feasibility. While, Zaghouani et al. [11] proposed a biometric scheme for patient authentication system in healthcare and hospitals as well, and the used technique was based on using Linear Prediction Coding (LPC) to hide the ECG sensitive data. Brown et al. [12] discussed the design of a prototype in which two commercial biometric devices were combined: the first device is used for reading the user's ECG signal, while the second one measures the user's pulses. Moreover, Arteaga-Falconi et al. [13] proposed a mobile biometric authentication algorithm based on ECG, and this algorithm needs 4 seconds of signal acquisition to identify the unknown individuals. Furthermore, Tantinger et al. [14] proposed an algorithm that is capable of segmenting individual heart beats from a single ECG channel. And, Yarong and Gang [15] represented a human identification scheme based on frequency-domain features that are extracted from the ECG signal; they used Fourier Transform (FT) to get the required features, which contain the signal slope, the harmonics number, the magnitude gap, and the frequency energy to the total energy ratio. While, Tantawi et al. [16] proposed a method for authenticating based on the discrete Wavelet feature extraction for an ECG based biometric system; for that, the RR intervals are processed using the discrete bi-orthogonal Wavelet in the coefficient structure, where the non-informative coefficient is excluded to reduce its structure, and then fed to the radial basis Neural Network (NN) for the classification process.

## III. DISCRETE COSINE TRANSFORM (DCT)

The DCT has a valuable property for applications requiring data processing; it can often rebuild a sequence precisely from only a few DCT coefficients. The DCT termed as ($y$) of a signal ($x$) is given by as in (1).

$$y(k) = \begin{cases} \frac{1}{\sqrt{N}} \sum_{n=0}^{N-1} x(n) & k = 0 \\ \sqrt{\frac{2}{N}} \sum_{n=0}^{N-1} x(n) \cos\left(\frac{\pi(2n-1)k}{2N}\right) & k = 1..,N-1 \end{cases} \quad (1)$$

Note that, $x$ and $y$ have the same sizes of $N$ samples. Like other transforms, the DCT also offers energy compaction. Moreover, the cosine bases of the DCT are orthogonal to each other [17].

## IV. CORRELATION COEFFICIENT

The correlation coefficient measuring method is widely used in the ECG authentication systems. It gives a statistical indicator that reflects the linear relationship between two variables. By calculating the correlation between two features groups, it can be determined whether the two groups belong to the same class or not. Consider the two datasets $K_i$ and $L_i$, and the correlation coefficient ($cf$) between them is defined as in (2) [18].

$$cf = \frac{\sum_i (K_i(L_i - \bar{L}) - \bar{K}(L_i - \bar{L}))}{\sqrt{(\sum_i (K_i - \bar{K})^2)(\sum_i (L_i - \bar{L})^2)}} \qquad (2)$$

Where,

$\bar{K}$ and $\bar{L}$ are the means of the datasets $K_i$ and $L_i$.

## V. METHODOLOGY

The block diagram of the ECG authentication proposed system is illustrated as in Fig. 3. It consists of a remote or a mobile side and a fixed (server) side. This structure gives the ability to combine multiple nodes with a single server, and this structure, in turn, increases the capability and reduces the whole cost as well. The mobile side consists of an ECG module for acquiring an ECG signal, a 16-bit high precision ADC, and a single-board computer (Raspberry PI 3 that operates by a customized Linux operating system. Moreover, the Raspberry PI 3 has the capability of operating the required algorithms that are based on Octave free license software. While, the server side contains a web server that manages the system database, which is the SQLite database.

And, Fig. 4 demonstrates the flowchart of the proposed system. The main concept of this system is to compare the features that extracted from the DCT coefficients for the first RR interval, which is saved in the database that hosted in server side, with next three consecutive intervals by using the cross-correlation technique. So, if the correlated result is more than 95% (Significant P-Value range) [19] for all durations, a person is evaluated to have an authorized access; otherwise, s/he is evaluated as an unauthorized person. And, this process is limited to 10 seconds, then the whole process will restart.

In this study, the MIT-BIH database for normal and abnormal records is used beside the real-time acquiring data for system validation and evaluation. Moreover, the main algorithm is distributed between the remote and the fixed sides. It will give a high flexibility and reduce the hardware cost.

Thus, the preprocessing stages, RR detection, and features extraction are done by using the DCT performed in Raspberry PI 3 as the remote side; while the web-based database and SQLite management application are hosted on the system/server side. And, Fig. 5 shows the hardware prototype implementation of the remote side.

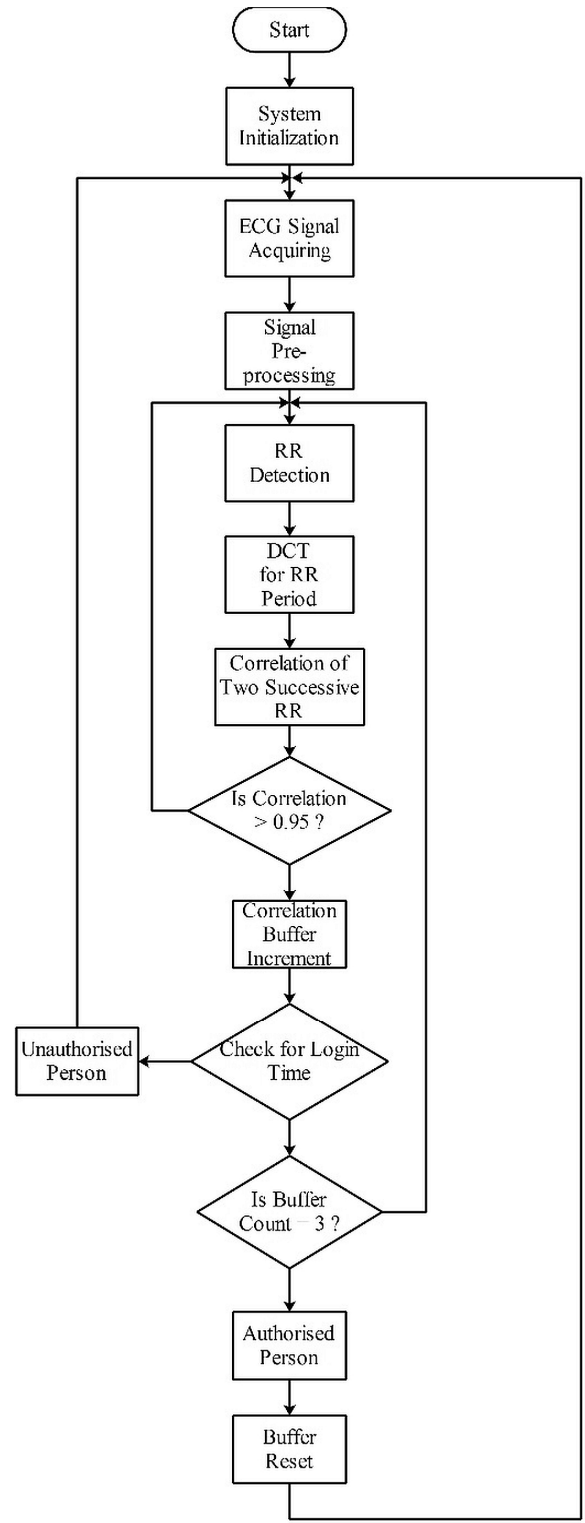

Fig. 4. The proposed system flowchart.

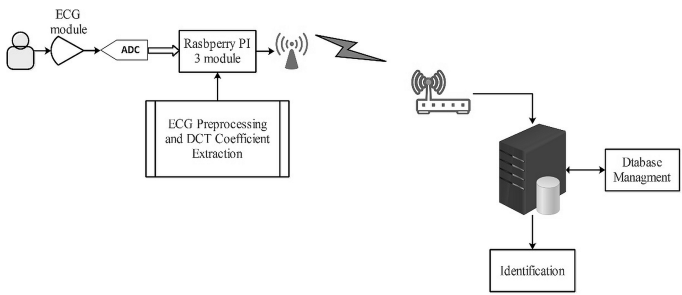

Fig. 3. The proposed ECG authentication system components.

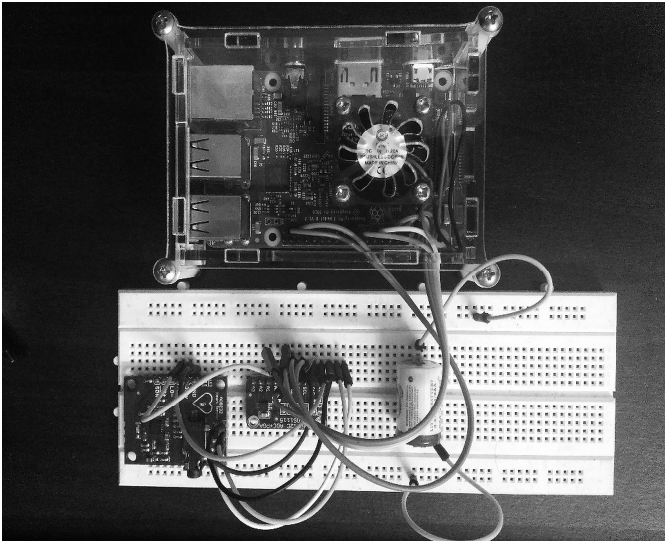

Fig. 5. The implemented mobile side (Raspberry PI 3).

## VI. RESULTS AND DISCUSSION

The proposed algorithm starts with measuring, acquiring, and preprocessing (filtering the unwanted signal components) of the ECG signal. The Raspberry PI 3 executes the acquiring software, where it controls the ADC that digitizes the analog ECG signal as well as the rest of processing algorithm parts. Fig. 6 demonstrates the web-based acquiring interface and the login page, where it is hosted and tested on Amazon Web Services (AWS).

Fig. 7 shows the acquired ECG signal with its RR detection periods. Note that, the first RR or fiducial mark is saved in the system's database and it will fetch during the authentication checking process.

The single RR intervals that used for features extraction are illustrated as in Fig. 8 (I), while the DCT extracted features results are demonstrated as in Fig. 8 (II). These results are used as unique marks for the authentication detection process. These features can give the specialized points that differ from a person to another. And, the total execution time for each checking varies between 1.02 to 1.21 seconds.

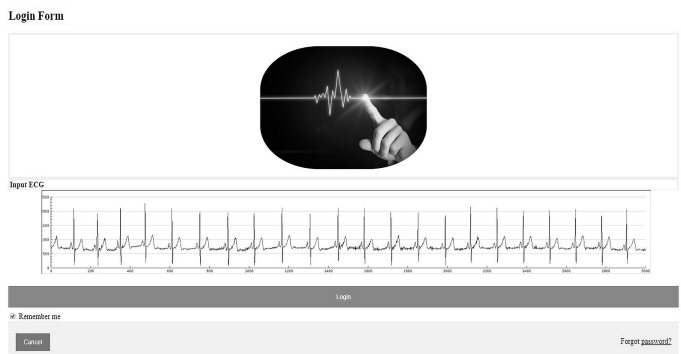

Fig. 6. Login web-based interface (server side).

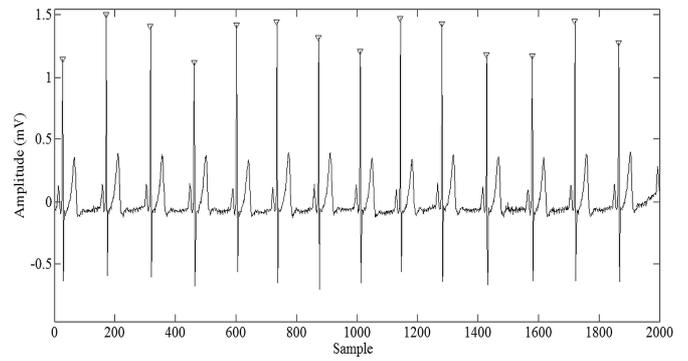

Fig. 7. The detected ECG RR intervals.

Table 1 states the result of 15 different subjects that used in this proposed system, and these subjects are taken from the measured ECG signals and from MIT-BIH database. The results demonstrate the correlation of the fiducial point (RR) intervals for three successive periods with the predefined and saved interval for a certain person. The correlation results that justified according to p-value show an accuracy rate of 97.78%, which is a promising outcome that shows the robustness of the proposed authentication system.

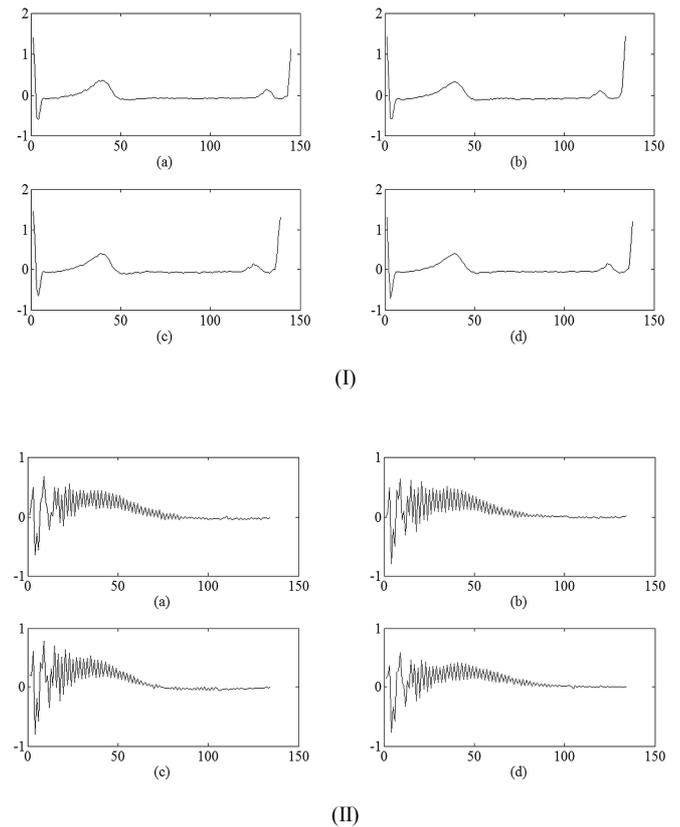

Fig. 8. Single beat for:

(I): (a) first RR, (b) second RR, (c) third RR, and (d) fourth RR intervals;

(II): (a) first RR, (b) second RR, (c) third RR, and (d) fourth RR features.

TABLE 1
EXTRACTED FEATURES CORRELATION RESULTS

| Obj. | Correlation/b | Obj. | Correlation/b | Obj. | Correlation/b |
|---|---|---|---|---|---|
| #1 | 96.13 | #6 | 97.37 | #11 | 88.05 |
|    | 97.37 |    | 98.02 |    | 95.77 |
|    | 97.34 |    | 96.67 |    | 96.85 |
| #2 | 93.97 | #7 | 96.76 | #12 | 97.96 |
|    | 98.36 |    | 98.92 |    | 99.37 |
|    | 98.89 |    | 95.99 |    | 97.59 |
| #3 | 99.46 | #8 | 97.33 | #13 | 95.88 |
|    | 97.95 |    | 97.74 |    | 95.26 |
|    | 98.27 |    | 98.55 |    | 91.47 |
| #4 | 95.99 | #9 | 92.12 | #14 | 95.21 |
|    | 96.26 |    | 98.96 |    | 99.78 |
|    | 94.74 |    | 95.95 |    | 96.81 |
| #5 | 99.17 | #10 | 99.54 | #15 | 95.36 |
|    | 98.87 |    | 96.35 |    | 98.22 |
|    | 98.66 |    | 95.89 |    | 99.02 |

## VII. CONCLUSION

Many ECG authentication researches are only discussed the accuracy issue and miscarried the real-time authentication process capability, accessing different points, and the real-time login issue. For that, the IoT real-time authentication system based on RR Discrete Cosine Transform features extraction is proposed in this study. This system uses the correlation of pre-defined RR features that stored in the system's database and three consecutive RR intervals from the acquired signal. And, the obtained accuracy rate of 97.78% at processing time around 1.21 seconds shows the dependability and the stability of the proposed system. For further development, this research study needs more data, especially from arrhythmia and ischemia cases, to discuss the different probabilities regarding the authentication detections.